\RequirePackage{fix-cm}
\documentclass[smallextended]{svjour3}       
\smartqed  
\usepackage{graphicx}
\usepackage{caption}
\usepackage{subcaption}
\usepackage[flushleft]{threeparttable}
\usepackage{geometry}
\begin{document}

\title{Protein Classification using Machine Learning and Statistical Techniques: A Comparative Analysis
}


\author{Chhote Lal Prasad Gupta \and
        Anand Bihari \and Sudhakar Tripathi 
}


\institute{C. L. P. Gupta \at
              Department of Computer Science \& Engineering, \\ Dr. A. P. J. Abdul Kalam Technical University, Lucknow, Uttar Pradesh. 
              \email{clpgupta@gmail.com}           
           \and
           A. Bihari \at
              Department of Computer Science \& Engineering, \\National Institute of Technology Patna
              \email{csanandk@gmail.com, anand.cse15@nitp.ac.in}
            \and
            S. Tripathi \at
            Department of Informtion Technology, \\
            R.E. C. Ambedkar Nagar, U. P. 
            \email{p.stripathi@gmail.com}
}

\date{Received: date / Accepted: date}

\maketitle

\begin{abstract}
In recent era prediction of enzyme class from an unknown protein is one of the challenging tasks in bioinformatics. Day to day the number of proteins is increases as result the prediction of enzyme class gives a new opportunity to bioinformatics scholars. The prime objective of this article is to implement the machine learning classification technique for feature selection and predictions also find out an appropriate classification technique for function prediction. In this article the seven different classification technique like CRT, QUEST, CHAID, C5.0, ANN (Artificial Neural Network), SVM and Bayesian has been implemented on 4368 protein data that has been extracted from UniprotKB databank and categories into six different class. The proteins data is high dimensional sequence data and contain a maximum of 48 features.To manipulate the high dimensional sequential protein data with different classification technique, the SPSS has been used as an experimental tool. Different classification techniques give different results for every model and shows that the data are imbalanced for class C4, C5 and C6. The imbalanced data affect the performance of model.  In these three classes the precision and recall value is very less or negligible. The experimental results highlight that the C5.0 classification technique accuracy is more suited for protein feature classification and predictions. The C5.0 classification technique gives 95.56\% accuracy and also gives high precision  and recall value. Finally, we conclude that the features that is selected can be used for function prediction. 

\end{abstract}

\section{Introduction}
\label{intro}
Protein is a sequence of amino acids binding with the peptide bond that play an important role in maintaining the life \cite{1}. Generally, it works in cells and help in function and regulation of tissues and organs of body \cite{2,3,4,5,5a}  The sequence of amino acids determines the function and structure of Protein. Mainly protein has three different structures Primary, Secondary and Tertiary. Protein function predictions play a key role to understand the diseases and a therapy to cure from diseases. Apart from experimental analysis several other approaches have been designed and implemented for protein function prediction like sequence similarity \cite{6}, clustering \cite{7}, protein interaction \cite{8} and many more \cite{9}. Several studies have been conducted to predict the function. Basically, the protein function prediction can be done by using sequence similarity, structure similarity or both. These method takes a lot of resources and computation time to predict the functions \cite{10}.  

From the last decades, it seems that the computational technique such as machine learning classification technique has been used for functional discovery and prediction of proteins that reduces the computation cost as well as human effort and increase the accuracy of function predication \cite{11,12,13,14,15}. \cite{16} proposed a new method for enzyme class prediction from protein and said that the overall accuracy of the proposed model is 35\%. Further \cite{17} used Bayesian classification technique and shows that the accuracy is improved and it is 45\%. \cite{11} used the machine learning approach for function prediction and used total 484 features for function prediction. In this article author used random forest technique and claimed that the accuracy is about 94.23\%. Apart from these researches several other researches have been conducted for protein function prediction with the help of machine learning technique such as SVM, ANN and Decision Tree and many researches indicates that the SVM gives better results than other classification technique \cite{10}.  \cite{18} used the SVM for protein function predictions. \cite{19} used SVM for finding DNA-binding sites. \cite{20} used the random forest to predict DNA-binding with the help of DNA-binding proteins and amino acid. In this article author mentioned that the hybrid approach of features selection may give better results and improve the accuracy of model. \cite{21} used the SVM and Nearest Neighbor classification technique for function prediction and claimed that the accuracy has been reached to 93.4\%.  Most of the research has used SVM as a classification technique to predict the function but this classification technique works efficiently for non-liner high dimensional data. The nature of the protein data is non-linear and have high dimension, but in many cases, it seems that the several features value are missing and SVM did not account this one and by using only one classification technique, we cannot say that the particular classification technique is suited for classification and predictions. To cover the missing data along with all possible data, in this article we made a comparative analysis of seven different classification technique like CRT, QUEST, CHAID, C5.0, ANN, SVM and Bayesian and found that the accuracy is above of 86 for all models and C5.0 gives highest accuracy above of 86\%. The C5.0 classification technique is capable to handle non-linearly high dimensional data including missing value. Finally, we can conclude that the C5.0 is more suited in such type of data sets. 

This article gives a comparative analysis of seven different classification technique on 4386 number of sample data of proteins that are classified into 6 different classes. In this article, first we discuss the process of data extraction that is one of the important parts of classification and prediction, feature extraction process and performance evaluation metrics in section 2. Section 3 deals with the experimental results and analysis and finally section 4 concludes the article.

\section{Data and Methods}
\label{sec:1}
In this section we have discussed data and important machine learning and statistical techniques that are implemented for protein classification.  
\subsection{Data collection}
\label{sec:2}
To predict the functional behavior of proteins from a wide range of different proteins a significant assignment.To do this, the UniProtKB (UniProtKnowledgebase) \cite{22} has been widely used. The data of UniProtKB is classified into two categories i.e. reviewed and un-reviewed. The reviewed dataset known as Swiss-Prot and contains 557,992 proteins. TrEMBL is un-reviewed dataset and contains 120,243,849 proteins.Generally, the reviewed dataset has been used for protein function prediction, because the data are reviewed and corrected by the user forums and communities \cite{23}. The Swiss-Prot dataset contains several organisms. In this study, we have considered only human organism and extracted the protein data of human’s enzyme class for protein classification. This data set contains total 4,368 no ofsamples of six different enzyme classes such as Oxidoreductases, Transferases, Hydrolases, Lyases, Isomerases, and Ligases (shown in Table \ref{tab:Protein Data Description}). This data set of proteins is used in experimental analysis and for finding best suitable features for function prediction.
\begin{center}
\begin{table}[h!]
	\renewcommand{\arraystretch}{1.2}
	\centering
	\caption{Protein Data Description}
	\label{tab:Protein Data Description}
	
	\begin{tabular}{lll}		\hline
		\textbf{Class}	&	\textbf{Name}	&	\textbf{No. of Proteins}	\\\hline
		EC1	&	Oxidoreductases	&	562	\\
		EC2	&	Transferases	&	1790	\\
		EC3	&	Hydrolases	&	1629	\\
		EC4	&	Lyases	&	149	\\
		EC5	&	Isomerases	&	112	\\
		EC6	&	Ligases	&	126	\\
		
		\hline
	\end{tabular}
\end{table}
\end{center}

\subsection{Feature extraction}
Feature selection for protein function prediction is one of the major tasks, because every protein contains a huge amount of feature. In real world a protein has many features and some of the features having very less significance in function prediction. The prime objective of feature selection is to eliminate the less significant feature that helps in accurate prediction and classification \cite{24,25}. Although, the feature that are eliminated may provide additional information that may improve the classification and prediction but adds the additional cost in classification and model may give different result. Several filter and wrapper feature selection techniques are available to extract potential feature. The wrapper techniques used machine learning technique; however, the filter technique used manual filtering process of feature selection and it is faster than wrapper. 
In this study we have used filter technique and select 45 potential features, which are present in all six classes of enzyme of human’s protein (shown in table \ref{tab:Protein Data Description}). These features are valuable in protein function prediction. 

\subsection{Classification of protein in enzyme class}
The proteins data are classified into six enzyme class (table 1) and every protein contains a set of features. Several studies related to protein classification mentioned that some of the proteins are wrongly classified into different class which affect the accuracy of function prediction. To classify proteins, several classification techniques are available. We have used the following classification techniques for protein classification: (i) CRT, (ii) QUEST, (iii) CHAID, (iv) 	C5.0, (v) ANN, (vi) SVM and (vii) Bayesian. \\

\textbf{(i) CRT:} In this classification, first data are classified with the help of classification tree and the predictions are based on the regression tree \cite{26}.\\

\textbf{(ii) QUEST:} It is a tree based binary classification technique. It reduces the computation time than the others tree-based classification.  In this classification, the statistical test has been conducted to select an input field. It also separates the input selection and the splitting of trees \cite{27}.\\

\textbf{(iii) CHAID:} It is a tree-based model for classification and prediction of variables and also find the interaction between variables. It builds a non-binary tree by using   multiple regression. The main objective of the CHAID technique is to find how one variable affect the performance of other variables \cite{28}.\\

\textbf{(iv) C5.0:} It is an extension of ID3 algorithm of decision tree. It produces a binary tree with multiple branch. It deals with all possible data including the missing features. It is discrete and continuous in nature \cite{29}.  \\

\textbf{(v) ANN:} It is basically used to estimate the performance of biological networks. In this technique the learning process is based on adjustment of weight between connection of neurons and the output of the model is depends on the activation function \cite{30}.\\

\textbf{(vi) SVM:} It is one of the most influenced classification techniques based on statistical learning for classifications and prediction of data \cite{31,32,33,34,35} It deals with wide variety of classification problems including the non-linearly high dimensional problem. \\

\textbf{(vii) Bayesian:} It is a graph-based classification technique. In this technique, the node represents the set of variables and the edge between nodes represents the conditional dependency between nodes. It also used in classification of sequence data \cite{36,37}.\\

\subsection{Performance evaluation metrics}
To estimate the performance of above mentioned seven classifiers, the following evaluation metrics: (a) Accuracy (AC), (b) Sensitivity (ST), (c) Specificity (SP), (d) F-measure and (e) MCC (Matthew’s correlation coefficient) are used in this research \cite{38,39}, which are described as:
\begin{equation}
AC=\frac{(TP+TN)}{(TP+FN+FP+TN)}
\end{equation}

\begin{equation}
ST=\frac{TP}{(TP+FN)}
\end{equation}

\begin{equation}
SP=\frac{TN}{(FP+TN)}
\end{equation}

\begin{equation}  
F-measure=\frac{(2 \times PR  \times RC)}{(PR+RC)}
\end{equation}

\begin{equation}
MCC=\frac{((TP \times TN)- (FP \times FN))}{((TP+FP)(TP+FN)(TN+FP)(TN+FN))}
\end{equation}

Where TP, FP, TN and FN represent the total numbers of true positive, false positive, true negative and false negative of proteins respectively. Precision (PR) and Recall (RC) are equivalent to sensitivity and specificity respectively. 

\subsection{Experimental result and analysis}
In this section, our main objective of experimental analysis is to find a small set of features for function prediction. To do this, first we have extracted the human’s protein data with 45 features, after that we have used above mentioned seven classification techniques for function predictions. Different classification techniques select different feature for functions predictions. To predict the functions which are plays important role in the function prediction, we have used SPSS as an experimental tool.  Our experimental result show that some of the features are present in all the seven classification techniques. The list of selected features with their importance factor given in bracket for every model is shown in Table \ref{tab:List of features selected by different models with their importance factor}. 
\begin{center}
	\begin{table}[h!]
		\renewcommand{\arraystretch}{1.2}
		\centering
		\caption{List of features selected by different models with their importance factor}
		\label{tab:List of features selected by different models with their importance factor}
		
		\begin{tabular}{lp{2cm}p{11cm}}		\hline
			\textbf{Model}	&	\textbf{Total No. of selected features}	&	\textbf{ Selected Feature with importance factor}	\\\hline
		CRT 	&	27	&	Beta strand(0.00040), Glycosylation(0.0013), Topological domain(0.0013), Site(0.0013), Peptide(0.0013),  Signal peptide(0.0013), Disulfide bond(0.0013), Modified residue(0.0013), Mass(0.0013), Erroneous termination(0.0013), Natural variant(0.0013), Transmembrane(0.0013), Chain(0.0013), Turn(0.0013), Sequence conflict(0.0013), Erroneous initiation(0.0013), utagenesis(0.0013), Length(0.0013), Propeptide(0.0013), Compositional bias(0.0013), Region(0.0074), Helix(0.0121), Domain(0.0723), Binding site(0.0918), Active site(0.2334), Metal binding(0.2653), Nucleotide binding(0.292)	\\\hline
		QUEST	&	18	&	Nucleotide binding(0.00), Binding site(0.00), Domain(0.0025), Erroneous termination(0.0025), Coiled coil(0.0025), Beta strand(0.0025), Disulfide bond(0.0025), Chain(0.0025), Helix(0.0025), Turn(0.0025), Modified residue(0.0025), Region(0.0025), Mutagenesis(0.0025), Compositional bias(0.0025), Propeptide(0.0025), Transit peptide(0.2524), Metal binding(0.2754), Active site(0.4399)	\\\hline
		CHAID 	&	15	&	Length(0.0004), Turn(0.0065), Signal peptide(0.0067), Region(0.0077), Motif(0.0111), Binding site(0.0174), Mass(0.0185), Glycosylation(0.0254), Modified residue(0.0284), Domain(0.0466), Zinc finger(0.0572), Helix(0.0614), Nucleotide binding(0.1519), Metal binding(0.1829), Active site(0.3778)	\\\hline
		C5.0	&	37	&	Site(0.0), Signal peptide(0.0), Disulfide bond(0.0), Helix(0.0) Modified residue(0.0), Region(0.0),Domain(0.0), Beta strand(0.0), Repeat(0.0), Sequence conflict(0.0), Sequence caution(0.0), Length(0.0), Initiator methionine(0.0), Compositional bias(0.0), Topological domain(0.0003), Erroneous initiation(0.001), Calcium binding(0.0011), Mass spectrometry(0.0011), Natural variant(0.0015), Erroneous gene model prediction(0.0022), Non-standard residue(0.00410), Alternative sequence(0.0052), Motif(0.0054), Transmembrane(0.0055), Glycosylation(0.0077), Frameshift(0.008), Cross-link(0.0297), Mass(0.0321), Turn(0.0346), Mutagenesis(0.0427), Propeptide(0.078), Nucleotide binding(0.086), Metal binding(0.1098), Transit peptide(0.1101), Zinc finger(0.1145), Binding site(0.1275), Active site(0.192	\\\hline
		ANN	&	45	&	Peptide(0.0025), Coiled coil(0.0063), Calcium binding(0.0064), Erroneous translation(0.007), DNA binding(0.0077), Sequence conflict(0.009), Intramembrane(0.0095), Erroneous termination(0.01), Signal peptide(0.0107), Erroneous gene model prediction(0.0112), Compositional bias(0.0112), Alternative sequence(0.0114), Transmembrane(0.0129), Disulfide bond(0.0129), Mass(0.0133), Initiator methionine(0.0136), Length(0.0137), Domain(0.0139), Sequence caution(0.0153), Lipidation(0.0154), Glycosylation(0.0179), Topological domain(0.0186), Turn(0.0193), Mutagenesis(0.0195), Site(0.0208), Frameshift(0.0219), Motif(0.022), Erroneous initiation(0.0224), Helix(0.0252), Chain(0.0259), Natural variant(0.0262), Modified residue(0.0268), Propeptide(0.0306), Region(0.0319), Beta strand(0.0339), Repeat(0.0351), Transit peptide(0.0364), Cross-link(0.0368), Binding site(0.0389), Mass spectrometry(0.0397), Non-standard residue(0.0423), Zinc finger(0.0438), Active site(0.047), Metal binding(0.05), Nucleotide binding(0.0533)	\\\hline
		SVM	&	45	&	Active site, Alternative sequence, Beta strand, Binding site, Calcium binding, Chain, Coiled coil, Compositional bias, Cross-link, DNA binding, Disulfide bond, Domain, Erroneous gene model, prediction, Erroneous initiation, Erroneous termination, Erroneous translation, Frameshift, Glycosylation, Helix, Initiator methionine, Intramembrane, Length, Lipidation, Mass, Mass spectrometry, Metal binding, Modified residue, Motif, Mutagenesis, Natural variant, Non-standard residue, Nucleotide binding, Peptide, Propeptide, Region, Repeat, Sequence caution, Sequence conflict, Signal peptide, Site, Topological domain, Transit peptide, Transmembrane, Turn, and Zinc finger.	\\\hline
		Bayesian	&	45	&	Signal peptide(0.0108), Beta strand(0.0158), Helix(0.0163), Mutagenesis(0.0167), Mass spectrometry(0.0167), Glycosylation(0.0168), Intramembrane(0.0172), Domain(0.0172), Erroneous termination(0.0172), Erroneous translation(0.0172), Disulfide bond(0.0172), Modified residue(0.0172), Peptide(0.0172), Non-standard residue(0.0172), Mass(0.0172), Length(0.0172), Repeat(0.0173), Sequence conflict(0.0173), Cross-link(0.0174), Frameshift(0.0174), Chain(0.0175), Lipidation(0.018), Erroneous initiation(0.0181), Calcium binding(0.0187), Initiator methionine(0.019), Metal binding(0.0202), Coiled coil(0.021), DNA binding(0.0221), Region(0.0224), Natural variant(0.0226), Nucleotide binding(0.0231), Alternative sequence(0.0235), Topological domain(0.0237), Sequence caution(0.0241), Compositional bias(0.0245), Turn(0.0245), Transmembrane(0.0262), Zinc finger(0.0263), Site(0.0281), Binding site(0.0284), Active site(0.0335), Motif(0.0343), Propeptide(0.0344), Erroneous gene model prediction(0.0362), Transit peptide(0.0753)	\\\hline
					
		\end{tabular}
	\end{table}
\end{center}
In this table (Table \ref{tab:List of features selected by different models with their importance factor}) one can clearly see that the CHAID classification technique select minimum number of features, i.e, 15 and the ANN, SVM and Bayesian classification technique select all of the features. This table also highlight the importance factor of all selected features for all models and one can see that a small number of features have high importance in the all model excluding SVM. Feature with high impact for every model is shown in Table \ref{tab:Model with high impact value}.
\begin{center}
	\begin{table}[h!]
		\renewcommand{\arraystretch}{1.2}
		\centering
		\caption{Model with high impact value}
		\label{tab:Model with high impact value}
		
		\begin{tabular}{ll}		\hline
			\textbf{Model}	&	\textbf{Feature Selected with high impact value}	\\\hline
	
		CRT 	&	 Nucleotide binding(0.292)	\\
		QUEST	&	 Active site(0.4399)	\\
		CHAID 	&	Active site(0.3778)	\\
		C5.0	&	Active site(0.192	\\
		ANN	&	Nucleotide binding(0.0533)	\\
		SVM	&	All	\\
		Bayesian	&	Transit peptide(0.0753)	\\\hline
		
		\end{tabular}
	\end{table}
\end{center}

From Table \ref{tab:Model with high impact value} one can conclude that the Active site feature have high importance is QUEST, CHAID and C5.0,  Nucleotide binding having high impact in CRT and ANN. Finally, we can conclude that the Active site features plays a significance role in feature selection.  In SVM all features have equal importance in classification. To find the optimal set of features that affect the classification and prediction, first we have analysed the impact of most influential features of every model and validated with the validation technique. 

\textbf{(i)	CRT based classification}\\
To find the optimal set of features that affect the function predictions, we have used CRT classification technique. In this classification, first data are classified with the help of classification tree and the predictions are based on the regression tree. The CRT classification is implemented on total 4,368 no of proteins of six different enzyme classes with 45 different features. This classification technique classified total 27features out of 45 are predicted for function predictions. The result of the predicted functions and the performance of model on our data-set is shown in Table \ref{tab:Performance result of the CRT model}.

\begin{center}
	\begin{table}[h!]
		\renewcommand{\arraystretch}{1.2}
		\centering
		\caption{Performance result of the CRT model for all classes}
		\label{tab:Performance result of the CRT model}
		
		\begin{tabular}{lllllll}		\hline
			\textbf{}	&	\textbf{C1}	&	\textbf{C2}&\textbf{C3} &\textbf{C4} & \textbf{C5} &\textbf{C6}	\\\hline
		True positive (TP)	&	228	&	1482	&	1149	&	0	&	0	&	0	\\
		True negative (TN)	&	3632	&	1749	&	2233	&	4219	&	4256	&	4242	\\
		False positive (FP)	&	174	&	829	&	506	&	0	&	0	&	0	\\
		False negative (FN)	&	334	&	308	&	480	&	149	&	112	&	126	\\
		&		&		&	&		&		&		\\
		Accuracy	&	88.36996337	&	73.96978022	&	77.42674	&	96.58883	&	97.4359	&	97.11538	\\
		Sensitivity	&	0.40569395	&	0.827932961	&	0.705341	&	0	&	0	&	0	\\
		Specificity	&	0.954282712	&	0.678432894	&	0.815261	&	1	&	1	&	1	\\
		Precision	&	0.567164179	&	0.641280831	&	0.69426	&	\#DIV/0!	&	\#DIV/0!	&	\#DIV/0!	\\
		Recall	&	0.40569395	&	0.827932961	&	0.705341	&	0	&	0	&	0	\\
		F-measure	&	0.473029046	&	0.722750549	&	0.699756	&	\#DIV/0!	&	\#DIV/0!	&	\#DIV/0!	\\
		MCC	&	0.416953394	&	0.498902024	&	0.518965	&	\#DIV/0!	&	\#DIV/0!	&	\#DIV/0!	\\
			
			\hline
		\end{tabular}
	\end{table}
\end{center}

\textbf{(ii)	QUEST: }\\
It is a tree based binary classification technique. It reduces the computation time than the others tree-based classification.  In this classification, the statistical test has been conducted to select an input field. It also separates the input selection and the splitting of trees. To find the optimal set of features, the QUEST has been implanted by using SPSS tool on above mentioned dataset. This classification techniques classified total 18 features for predictions. The result of the performance of this model is shown in Table \ref{tab:Performance result of the QUEST model}.
\begin{center}
	\begin{table}[h!]
		\renewcommand{\arraystretch}{1.2}
		\centering
		\caption{Performance result of the QUEST model for all classes}
		\label{tab:Performance result of the QUEST model}
		
		\begin{tabular}{lllllll}		\hline
			\textbf{}	&	\textbf{C1}	&	\textbf{C2}&\textbf{C3} &\textbf{C4} & \textbf{C5} &\textbf{C6}	\\\hline
			True positive (TP)	&	0	&	1544	&	1148	&	0		&0	&	0	\\
			True negative (TN)	&	3806	&	1483	&	2158	&	4219	&	4256	&	4242	\\
			False positive (FP)	&	0	&	1095	&	581	&	0	&	0	&	0	\\
			Fasle negative (FN)	&	562	&	246	&	481	&	149	&	112	&	126	\\
			&		&	&		&		&		&		\\
			Accuracy	&	87.13369963	&	69.29945	&	75.68681	&	96.58883	&	97.4359	&	97.11538	\\
			Sensitivity	&	0	&	0.86257	&	0.704727	&	0	&	0	&	0	\\
			Specificity	&	1	&	0.575252	&	0.787879	&	1	&	1	&	1	\\
			Precision	&	\#DIV/0!	&	0.58507	&	0.663968	&	\#DIV/0!	&	\#DIV/0!	&	\#DIV/0!	\\
			Recall	&	0	&	0.86257	&	0.704727	&	0	&	0	&	0	\\
			F-measure	&	\#DIV/0!	&	0.697223	&	0.68374	&	\#DIV/0!	&	\#DIV/0!	&	\#DIV/0!	\\
			MCC	&	\#DIV/0!	&	0.4403	&	0.487123	&	\#DIV/0!	&	\#DIV/0!	&	\#DIV/0!	\\
		
			\hline
		\end{tabular}
	\end{table}
\end{center}

\textbf{iii) CHAID (Chi-square Automatic Interaction Detection)	}\\
It is a tree-based model for classification and prediction of variables and also find the interaction between variables. It builds a non-binary tree by using   multiple regression. The main objective of the CHAID technique is to find how one variable affect the performance of other variables. The CHAID classification is implemented on total 4,368 no of proteins of six different enzyme classes with 45 different features. This classification technique classified total 15 features out of 45 are predicted for function predictions. The result of the predicted functions and the performance of model on our data-set is shown in Table \ref{tab:Performance result of the CHAID model}.

\begin{center}
	\begin{table}[h!]
		\renewcommand{\arraystretch}{1.2}
		\centering
		\caption{Performance result of the CHAID model for all classes}
		\label{tab:Performance result of the CHAID model}
		
		\begin{tabular}{lllllll}		\hline
			\textbf{}	&	\textbf{C1}	&	\textbf{C2}&\textbf{C3} &\textbf{C4} & \textbf{C5} &\textbf{C6}	\\\hline
		True positive (TP)	&	241	&	1466	&	1260	&	0	&	0	&	18	\\
		True negative (TN)	&	3646	&	1878	&	2257	&	4219	&	4256	&	4201	\\
		False positive (FP)	&	160	&	700	&	482	&	0	&	0	&	41	\\
		Fasle negative (FN)	&	321	&	324	&	369	&	149	&	112	&	108	\\
		&		&		&		&		&		&		\\
		Accuracy	&	88.9881	&	76.55678	&	80.5174	&	96.58883	&	97.4359	&	96.58883	\\
		Sensitivity	&	0.428826	&	0.818994	&	0.773481	&	0	&	0	&	0.142857	\\
		Specificity	&	0.957961	&	0.728472	&	0.824023	&	1	&	1	&	0.990335	\\
		Precision	&	0.600998	&	0.676824	&	0.723307	&	\#DIV/0!	&	\#DIV/0!	&	0.305085	\\
		Recall	&	0.428826	&	0.818994	&	0.773481	&	0	&	0	&	0.142857	\\
		F-measure	&	0.500519	&	0.741153	&	0.747553	&	\#DIV/0!	&	\#DIV/0!	&	0.194595	\\
		MCC	&	0.448509	&	0.538502	&	0.5901	&	\#DIV/0!	&	\#DIV/0!	&	0.193123	\\

			\hline
		\end{tabular}
	\end{table}
\end{center}
\textbf{(iv)  SVM:}\\ 
It is one of the most influenced classification techniques based on statistical learning for classifications and prediction of data [7,8,9]. It deals with wide variety of classification problems including the non-linearly high dimensional problem. It provides more efficient predication than others classification problem. But it requires a set of key parameters. We have implemented this technique on our data sets and we found that it takes all 45 features as an input and predict total 45 number of features for function prediction.The result of performance of the SVM is shown in Table \ref{tab:Performance result of the SVM model}
\begin{center}
	\begin{table}[h!]
		\renewcommand{\arraystretch}{1.2}
		\centering
		\caption{Performance result of the SVM model for all classes}
		\label{tab:Performance result of the SVM model}
		
		\begin{tabular}{lllllll}		\hline
			\textbf{}	&	\textbf{C1}	&	\textbf{C2}&\textbf{C3} &\textbf{C4} & \textbf{C5} &\textbf{C6}	\\\hline
		True positive (TP)	&	164	&	1495	&	1113	&	19	&	15	&	32	\\
		True negative (TN)	&	3646	&	1694	&	2303	&	4209	&	4242	&	4215	\\
		False positive (FP)	&	160	&	884	&	436	&	10	&	14	&	27	\\
		False negative (FN)	&	398	&	296	&	516	&	129	&	97	&	94	\\
		&		&		&		&		&		&		\\
		Accuracy	&	87.22527	&	73.00824	&	78.20513	&	96.79487	&	97.45879	&	97.22985	\\
		Sensitivity	&	0.291815	&	0.834729	&	0.683241	&	0.128378	&	0.133929	&	0.253968	\\
		Specificity	&	0.957961	&	0.657099	&	0.840818	&	0.99763	&	0.996711	&	0.993635	\\
		Precision	&	0.506173	&	0.628415	&	0.718528	&	0.655172	&	0.517241	&	0.542373	\\
		Recall	&	0.291815	&	0.834729	&	0.683241	&	0.128378	&	0.133929	&	0.253968	\\
		F-measure	&	0.370203	&	0.717026	&	0.700441	&	0.214689	&	0.212766	&	0.345946	\\
		MCC	&	0.319136	&	0.485712	&	0.529741	&	0.280731	&	0.254267	&	0.359015	\\
				\hline
		\end{tabular}
	\end{table}
\end{center}

\textbf{(V) C5.0}\\
It is an extension of ID3 algorithm of decision tree. It produces a binary tree with multiple branch. It deals with all possible data including the missing features. It is discrete and continuous in nature.  In protein function predictions several data have missing some features value. This classification technique is more suited in this case. We have implemented this technique on our data set and found that it selects maximum number of features i.e. 37 in functions predictions. The performance of this model is shown in Table \ref{tab:Performance result of the C5.0 model}.
\begin{center}
	\begin{table}[h!]
		\renewcommand{\arraystretch}{1.2}
		\centering
		\caption{Performance result of the C5.0 model for all classes}
		\label{tab:Performance result of the C5.0 model}
		
		\begin{tabular}{lllllll}		\hline
			\textbf{}	&	\textbf{C1}	&	\textbf{C2}&\textbf{C3} &\textbf{C4} & \textbf{C5} &\textbf{C6}	\\\hline
			True positive (TP)	&	452	&	1653	&	1463	&	83	&	58	&	69	\\
			True negative (TN)	&	3722	&	2332	&	2530	&	4192	&	4246	&	4228	\\
			False positive (FP)	&	84	&	246	&	209	&	27	&	10	&	14	\\
			Fasle negative (FN)	&	110	&	137	&	166	&	66	&	52	&	57	\\
			&		&		&		&		&		&		\\
			Accuracy	&	95.55861	&	91.23168	&	91.41484	&	97.87088	&	98.5348	&	98.37454	\\
			Sensitivity	&	0.80427	&	0.923464	&	0.898097	&	0.557047	&	0.527273	&	0.547619	\\
			Specificity	&	0.97793	&	0.904577	&	0.923695	&	0.9936	&	0.99765	&	0.9967	\\
			Precision	&	0.843284	&	0.870458	&	0.875	&	0.754545	&	0.852941	&	0.831325	\\
			Recall	&	0.80427	&	0.923464	&	0.898097	&	0.557047	&	0.527273	&	0.547619	\\
			F-measure		&0.823315	&	0.896178	&	0.886398	&	0.640927		&0.651685	&	0.660287	\\
			MCC	&	0.798225	&	0.821479	&	0.817599	&	0.637929		&0.664363		&0.667282	\\
			
			\hline
		\end{tabular}
	\end{table}
\end{center}
\textbf{(Vi) Bayesian Network }\\
It is a graph-based classification technique. In this technique, the node represents the set of variables and the edge between nodes represents the conditional dependency between nodes. It also used in classification of sequence data. Our protein data is a sequence data, so that this technique is useful in classification and prediction of data. We have implemented this technique on our dataset and found that it selects total 45 features in classification and predict 45 number of features for function prediction. The performance of Bayes network model is shown in Table \ref{tab:Performance result of the Bayesian model}.
\begin{center}
	\begin{table}[h!]
		\renewcommand{\arraystretch}{1.2}
		\centering
		\caption{Performance result of the Bayesian model for all classes}
		\label{tab:Performance result of the Bayesian model}
		
		\begin{tabular}{lllllll}		\hline
			\textbf{}	&	\textbf{C1}	&	\textbf{C2}&\textbf{C3} &\textbf{C4} & \textbf{C5} &\textbf{C6}	\\\hline
		True positive (TP)	&	124	&	1557	&	931	&	19	&	20	&	18	\\
		True negative (TN)	&	3695	&	1332	&	2432	&	4200	&	4252	&	4230	\\
		False positive (FP)	&	111	&	1246	&	307	&	19	&	4	&	12	\\
		Fasle negative (FN)	&	438	&	233	&	698	&	130	&	92	&	108	\\
		&		&		&		&		&		&		\\
		Accuracy	&	87.43132	&	66.14011	&	76.99176	&	96.58883	&	97.8022	&	97.25275	\\
		Sensitivity	&	0.220641	&	0.869832	&	0.571516	&	0.127517	&	0.178571	&	0.142857	\\
		Specificity	&	0.970836	&	0.51668	&	0.887915	&	0.995497	&	0.99906	&	0.997171	\\
		Precision	&	0.52766	&	0.555476	&	0.752019	&	0.5	&	0.833333	&	0.6	\\
		Recall	&	0.220641	&	0.869832	&	0.571516	&	0.127517	&	0.178571	&	0.142857	\\
		F-measure	&	0.311167	&	0.677988	&	0.649459	&	0.203209	&	0.294118	&	0.230769	\\
		MCC	&	0.284152	&	0.396426	&	0.492998	&	0.240444	&	0.379821	&	0.283779	\\

			\hline
		\end{tabular}
	\end{table}
\end{center}

\textbf{(Vii) ANN}\\
It is basically used to estimate the performance of biological networks. In this technique the learning process is based on adjustment of weight between connection of neurons.  The activity of the neural network is referred as the liner combination of set of neurons and the output of the model is depends on the activation function. This technique also helps in classification of sequence-based data.  We have implemented neural network model on our data set and found that it has taken all 45 features as an input and predict all 45 features for function prediction. The performance of the Neural network is shown in Table \ref{tab:Performance result of the ANN model}. 
\begin{center}
	\begin{table}[h!]
		\renewcommand{\arraystretch}{1.2}
		\centering
		\caption{Performance result of the ANN model for all classes}
		\label{tab:Performance result of the ANN model}
		
		\begin{tabular}{lllllll}		\hline
			\textbf{}	&	\textbf{C1}	&	\textbf{C2}&\textbf{C3} &\textbf{C4} & \textbf{C5} &\textbf{C6}	\\\hline
		True positive (TP)	&	156	&	1471	&	1248	&	0	&	0	&	0	\\
		True negative (TN)	&	3566	&	1805	&	2259	&	4219	&	4256	&	4242	\\
		False positive (FP)	&	240	&	773	&	480	&	0	&	0	&	0	\\
		Fasle negative (FN)	&	406	&	319	&	381	&	149	&	112	&	126	\\
		&		&		&		&		&		&		\\
		Accuracy	&	85.21062	&	75	&	80.28846	&	96.58883	&	97.4359	&	97.11538	\\
		Sensitivity	&	0.27758	&	0.821788	&	0.766114	&	0	&	0	&	0	\\
		Specificity	&	0.936942	&	0.700155	&	0.824754	&	1	&	1	&	1	\\
		Precision	&	0.393939	&	0.655526	&	0.722222	&	\#DIV/0!	&	\#DIV/0!	&	\#DIV/0!	\\
		Recall	&	0.27758	&	0.821788	&	0.766114	&	0	&	0	&	0	\\
		F-measure	&	0.325678	&	0.729301	&	0.743521	&	\#DIV/0!	&	\#DIV/0!	&	\#DIV/0!	\\
		MCC	&	0.250162	&	0.513573	&	0.58435	&	\#DIV/0!	&	\#DIV/0!	&	\#DIV/0!	\\
				
			\hline
		\end{tabular}
	\end{table}
\vspace{-1.5cm}
\end{center}
\section{Analysis of result}
In this section our main objective is to find out the appropriate model for protein function predictions. To do this, we have made two different comparative analysis: (i) Model wise comparative analysis of all model for every classes and (ii) Class wise comparative analysis of all model. These two comparative analyses help in finding the appropriate model for function prediction and also helps to find out the appropriate model for every class. The class wise comparative analysis of performance of CRT based model is shown in Table \ref{tab:A Comparative analysis of performance of CRT for all class }. 
\begin{center}
	\begin{table}[h!]
		\renewcommand{\arraystretch}{1.2}
		\centering
		\caption{A Comparative analysis of performance of CRT for all class }
		\label{tab:A Comparative analysis of performance of CRT for all class }
		
		\begin{tabular}{llllllll}		\hline
			\textbf{}	&	\textbf{Accuracy}	&	\textbf{Sensitivity}&\textbf{Specificity} &\textbf{Precision} & \textbf{Recall} &\textbf{F-measure}&\textbf{MCC}	\\\hline
		C1	&	88.36996337	&	0.40569395	&	0.954282712	&	0.567164179	&	0.40569395	&	0.473029046	&	0.416953394	\\
		C2	&	73.96978022	&	\textbf{0.827932961}	&	0.678432894	&	0.641280831	&	\textbf{0.827932961}	&	\textbf{0.722750549}	&	0.498902024	\\
		C3	&	77.42673993	&	0.7053407	&	0.815261044	&	\textbf{0.694259819}	&	0.7053407	&	0.699756395	&	\textbf{0.518965253}	\\
		C4	&	96.58882784	&	0	&	\textbf{1}	&	\#DIV/0!	&	0	&	\#DIV/0!	&	\#DIV/0!	\\
		C5	&	\textbf{97.43589744}	&	0	&	\textbf{1}	&	\#DIV/0!	&	0	&	\#DIV/0!	&	\#DIV/0!	\\
		C6	&	97.11538462	&	0	&	\textbf{1}	&	\#DIV/0!	&	0	&	\#DIV/0!	&	\#DIV/0!	\\
		
			\hline
		\end{tabular}
	\end{table}
\end{center}
It can be clearly seen that the accuracy of class 5 is 97.4358, the sensitivity of class 2 is 0.8279, the specificity of the class 4, 5, \& 6 is 1, the Precision of class 3 is 0.6942, the recall is 0.8279 for class 2, the F-measures is 0.7227 for class 2 and the MCC is 0.5189 for class 3 is high as compare to other classes. If we consider the accuracy as a measure then this model works good for class 5 and class 6, because both of the classes’ accuracy have very less difference, but their precision and recall ration is zero. Therefore, we can say that this model is under-fitting in case of class C4, C5 and C6 and over-fitting in case of class C1, C2 and C3. If we consider class C1, C2 and C3, then it can be clearly seen that the class C1 accuracy is more than the other 2 classes, but their precision and recall value is relatively lesser than the class C2 and C3. If we consider the higher precision, recall and F-measures, then the performance of this model is good for class C2. If we consider MCC as an evaluation metrics, then class C3 is good. Finally, we can conclude that the CRT based model works efficiently for class C2. The class wise comparative analysis of performance of QUEST based model is shown in Table \ref{tab:A Comparative analysis of performance of QUEST for all class }. 
\begin{center}
	\begin{table}[h!]
		\renewcommand{\arraystretch}{1.2}
		\centering
		\caption{A Comparative analysis of performance of QUEST for all class }
		\label{tab:A Comparative analysis of performance of QUEST for all class }
		
		\begin{tabular}{llllllll}		\hline
			\textbf{}	&	\textbf{Accuracy}	&	\textbf{Sensitivity}&\textbf{Specificity} &\textbf{Precision} & \textbf{Recall} &\textbf{F-measure}&\textbf{MCC}	\\\hline
		C1	&	87.13369963	&	0	&	0	&	0	&	0	&	0	&	0	\\
		C2	&	69.29945055	&\textbf{0.862569832}	&	0.575252133	&	0.585070102	&	\textbf{0.862569832}	&	\textbf{0.697222849}	&	0.440299636	\\
		C3	&	75.68681319	&	0.704726826	&	0.787878788	&	\textbf{0.663967611}	&	0.704726826	&	0.683740322	&\textbf{0.487123099}	\\
		C4	&	96.58882784	&	0	&	\textbf{1}	&	\#DIV/0!	&	0	&	\#DIV/0!	&	\#DIV/0!	\\
		C5	&	\textbf{97.43589744}	&	0	&	\textbf{1}	&	\#DIV/0!	&	0	&	\#DIV/0!	&	\#DIV/0!	\\
		C6	&	97.11538462	&	0	&	\textbf{1}	&	\#DIV/0!	&	0	&	\#DIV/0!	&	\#DIV/0!	\\	
			\hline
		\end{tabular}
	\end{table}
\end{center}
Here it can be clearly seen that the performance of the QUEST model on this data set is degraded as compare to the CRT model. Model is underfitted in case of class C1, C4, C5, and C6 and overfitted in case of class C2 and C3, because the data are imbalanced.  This model works efficiently for class C5, when we consider high accuracy, in case of high sensitivity, recall or high F-measure, it works for class C2, and in case of high specificity, precision, or MCC, then model works efficiently for class C3. Finally, we can conclude that the QUEST model works efficiently on class C3 data and produce an average performance in every measure. The class wise comparative analysis of performance of CHAID based model is shown in Table \ref{tab:A Comparative analysis of performance of CHAID for all class }.
\begin{center}
	\begin{table}[h!]
		\renewcommand{\arraystretch}{1.2}
		\centering
		\caption{A Comparative analysis of performance of CHAID for all class }
		\label{tab:A Comparative analysis of performance of CHAID for all class }
		
		\begin{tabular}{llllllll}		\hline
			\textbf{}	&	\textbf{Accuracy}	&	\textbf{Sensitivity}&\textbf{Specificity} &\textbf{Precision} & \textbf{Recall} &\textbf{F-measure}&\textbf{MCC}	\\\hline
		C1	&	88.98809524	&	0.428825623	&	0.957961114	&	0.600997506	&	0.428825623	&	0.500519211	&	0.448508664	\\
		C2	&	76.55677656	&	\textbf{0.818994413}	&	0.728471683	&	0.676823638	&	\textbf{0.818994413}	&	0.741152679	&	0.538501997	\\
		C3	&	80.51739927	&	0.773480663	&	0.824023366	&	\textbf{0.723306544}	&	0.773480663	&	\textbf{0.747552655}	&	\textbf{0.59010047}	\\
		C4	&	96.58882784	&	0	&	\textbf{1}	&	\#DIV/0!	&	0	&	\#DIV/0!	&	\#DIV/0!	\\
		C5	&	\textbf{97.43589744}	&	0	&	\textbf{1}	&	\#DIV/0!	&	0	&	\#DIV/0!	&	\#DIV/0!	\\
		C6	&	96.58882784	&	0.142857143	&	0.990334748	&	0.305084746	&	0.142857143	&	0.194594595	&	0.193123061	\\
		
			\hline
		\end{tabular}
	\end{table}
\end{center}
In Table \ref{tab:A Comparative analysis of performance of CHAID for all class }, it can be clearly seen that the performance of CHAID is much better than the CRT and QUEST.  Here, the model is underfitted in case of Class C4 and C5 an overfitted in case of other classes. The CRT and QUEST model are underfitted in case of class C4 and C5 and also here it is underfitted. If we consider high accuracy, then this model works efficiently for, class C5, when consider only sensitivity, then model fitted for class C2, in case of high specificity model works efficiently for class C6, when we consider precision, F-measures or MCC, then model works efficiently for class C3 and in case we consider only recall then it works efficiently for class C2. Finally, we can conclude that this model works efficiently for class C3 and produce on average performance in every measure.  The class wise comparative analysis of performance of C5.0 based model is shown in Table \ref{tab:A Comparative analysis of performance of C5.0 for all class }.
\begin{center}
	\begin{table}[h!]
		\renewcommand{\arraystretch}{1.2}
		\centering
		\caption{A Comparative analysis of performance of C5.0 for all class }
		\label{tab:A Comparative analysis of performance of C5.0 for all class }
		
		\begin{tabular}{llllllll}		\hline
			\textbf{}	&	\textbf{Accuracy}	&	\textbf{Sensitivity}&\textbf{Specificity} &\textbf{Precision} & \textbf{Recall} &\textbf{F-measure}&\textbf{MCC}	\\\hline
		C1	&	95.55860806	&	0.804270463	&	0.977929585	&	0.843283582	&	0.804270463	&	0.823315118	&	0.798224848	\\
		C2	&	91.23168498	&	\textbf{0.923463687}	&	0.904577192	&	0.870458136	&	\textbf{0.923463687}	&	\textbf{0.896177826}	&	\textbf{0.821479485	}\\
		C3	&	91.41483516	&	0.898096992	&	0.923694779	&	\textbf{0.875}	&	0.898096992	&	0.886398061	&	0.817598839	\\
		C4	&	97.87087912	&	0.55704698	&	0.993600379	&	0.754545455	&	0.55704698	&	0.640926641	&	0.637928913	\\
		C5	&	\textbf{98.53479853}	&	0.527272727	&	\textbf{0.997650376}	&	0.852941176	&	0.527272727	&	0.651685393	&	0.664362603	\\
		C6	&	98.37454212	&	0.547619048	&	0.99669967	&	0.831325301	&	0.547619048	&	0.660287081	&	0.667282006	\\

			\hline
		\end{tabular}
	\end{table}
\end{center}
Table \ref{tab:A Comparative analysis of performance of C5.0 for all class } shows that the performance of C5.0 model is much better than the CRT, QUEST and CHAID. This model classify data appropriately and produce balanced result and the accuracy in all classes in above of 90\%. Similarly, if we consider the accuracy as a measure, then this model works efficiently for class C5, in case of sensitivity, it works efficiently for class C2, in case of specificity and precision, the model works efficiently for class C5 and C3 respectively. If we consider recall, F-measures or MCC, it works efficiently on class C2 data. If we consider all measures, then it works efficiently for class C2, because all measures value is above of 80\%.  The class wise comparative analysis of performance of ANN based model is shown in Table \ref{tab:A Comparative analysis of performance of ANN for all class }.
\begin{center}
	\begin{table}[h!]
		\renewcommand{\arraystretch}{1.2}
		\centering
		\caption{A Comparative analysis of performance of ANN for all class }
		\label{tab:A Comparative analysis of performance of ANN for all class }
		
		\begin{tabular}{llllllll}		\hline
			\textbf{}	&	\textbf{Accuracy}	&	\textbf{Sensitivity}&\textbf{Specificity} &\textbf{Precision} & \textbf{Recall} &\textbf{F-measure}&\textbf{MCC}	\\\hline
		C1	&	85.21062271	&	0.277580071	&	0.936941671	&	0.393939394	&	0.277580071	&	0.325678497	&	0.25016218	\\
		C2	&	75	&	\textbf{0.821787709}	&	0.700155159	&	0.655525847	&	\textbf{0.821787709}	&	0.729300942	&	0.513573088	\\
		C3	&	80.28846154	&	0.76611418	&	0.82475356	&	\textbf{0.722222222}	&	0.76611418	&	\textbf{0.743521001}	&	\textbf{0.584349942}	\\
		C4	&	96.58882784	&	0	&	\textbf{1}	&	\#DIV/0!	&	0	&	\#DIV/0!	&	\#DIV/0!	\\
		C5	&	\textbf{97.43589744}	&	0	&	\textbf{1}	&	\#DIV/0!	&	0	&	\#DIV/0!	&	\#DIV/0!	\\
		C6	&	97.11538462	&	0	&	\textbf{1}	&	\#DIV/0!	&	0	&	\#DIV/0!	&	\#DIV/0!	\\
				\hline
		\end{tabular}
	\vspace{-0.5cm}
	\end{table}
\end{center}
This model performs as similar to CRT and under-fitted for class C4, C5 and C6 and over-fitted for rest of the classes. As per accuracy, one can say that the model works efficiently for class C5 and produce similar specificity value for class C4, C5 and C6. If we consider high precision or recall value, it works efficiently for class C3 and C2 respectively. When we consider, F-measures or MCC as an evaluation metrics then the model works efficiently for class C3. Overall, we can say that the Neural network model works efficiently for class C3 and produce on average value for all measures above of 70\%. Table \ref{tab:A Comparative analysis of performance of SVM for all class } shows the class wise comparative analysis of performance of SVM based model.
\begin{center}
	\begin{table}[h!]
		\renewcommand{\arraystretch}{1.2}
		\centering
		\caption{A Comparative analysis of performance of SVM for all class }
		\label{tab:A Comparative analysis of performance of SVM for all class }
		
		\begin{tabular}{llllllll}		\hline
			\textbf{}	&	\textbf{Accuracy}	&	\textbf{Sensitivity}&\textbf{Specificity} &\textbf{Precision} & \textbf{Recall} &\textbf{F-measure}&\textbf{MCC}	\\\hline
		C1	&	87.22527473	&	0.291814947	&	0.957961114	&	0.50617284	&	0.291814947	&	0.37020316	&	0.319135622	\\
		C2	&	73.00824176	&	\textbf{0.834729202}	&	0.657098526	&	0.628415301	&	\textbf{0.834729202}	&\textbf{0.717026379}	&	0.485711627	\\
		C3	&	78.20512821	&	0.683241252	&	0.840817817	&	\textbf{0.718528083}	&	0.683241252	&	0.700440529	&	\textbf{0.529740958}	\\
		C4	&	96.79487179	&	0.128378378	&	\textbf{0.99762977}	&	0.655172414	&	0.128378378	&	0.214689266	&	0.280731068	\\
		C5	&	\textbf{97.45879121}	&	0.133928571	&	0.996710526	&	0.517241379	&	0.133928571	&	0.212765957	&	0.254266514	\\
		C6	&	97.22985348	&	0.253968254	&	0.993635078	&	0.542372881	&	0.253968254	&	0.345945946	&	0.3590152	\\
			\hline
		\end{tabular}
	\vspace{-0.25cm}
	\end{table}
\end{center}
It performs almost similar to C5.0 and can say that the successor model of CRT, QUEST, CHAID and Neural.  As similar to previous model, it works efficiently for class C5 when we consider accuracy as an evaluation metrics. If we consider either sensitivity or F-measures, it works efficiently for class C2. If we consider either precision, recall or MCC, it works efficiently for class C3. When we find the overall performance of this model in every measure, it works efficiently for class C3. The comparative analysis of performance of Bayesian model can see in Table \ref{tab:A Comparative analysis of performance of Bayesian for all class}.
\begin{center}
	\begin{table}[h!]
		\renewcommand{\arraystretch}{1.2}
		\centering
		\caption{A Comparative analysis of performance of Bayesian for all class }
		\label{tab:A Comparative analysis of performance of Bayesian for all class}
		
		\begin{tabular}{llllllll}		\hline
			\textbf{}	&	\textbf{Accuracy}	&	\textbf{Sensitivity}&\textbf{Specificity} &\textbf{Precision} & \textbf{Recall} &\textbf{F-measure}&\textbf{MCC}	\\\hline
		C1	&	87.43131868	&	0.220640569	&	0.970835523	&	0.527659574	&	0.220640569	&	0.311166876	&	0.284151839	\\
		C2	&	66.14010989	&	\textbf{0.869832402}	&	0.516679597	&	0.555476275	&	\textbf{0.869832402}	&	\textbf{0.677988243}	&	0.396426093	\\
		C3	&	76.99175824	&	0.571516268	&	0.887915298	&	0.752019386	&	0.571516268	&	0.649459365	&	\textbf{0.492997705}	\\
		C4	&	96.58882784	&	0.127516779	&	0.995496563	&	0.5	&	0.127516779	&	0.203208556	&	0.240444234	\\
		C5	&	\textbf{97.8021978}	&	0.178571429	&	\textbf{0.99906015}	&	\textbf{0.833333333}	&	0.178571429	&	0.294117647	&	0.379821433	\\
		C6	&	97.25274725	&	0.142857143	&	0.997171146	&	0.6	&	0.142857143	&	0.230769231	&	0.283779477	\\
		
			\hline
		\end{tabular}
	\vspace{-0.5cm}
	\end{table}
\end{center}
This model performs as similar to SVM and C5.0 and give balanced result.  In this table number in boldfaces indicates that the model performs efficiently for respective class. From the above discussion it can be clearly seen that the accuracy of all model is high for class C5 as compare to other classes and rest of the measures value is different for different class, however the precision and recall value for class C5 is relatively very low or negligible in every model. From the above analysis we can conclude that the almost all of the above classification technique work efficiently for class C3 excluding the CRT and Bayesian. The CRT and Bayesian classification technique work efficient for class C2. The experimental results highlight that the Class C4, C5 and C6 data are comparatively more imbalance than the class of C1, C2 and C3. Finally, we can conclude that the C5.0, SVM and Bayesian classification technique can be used for data classification and function prediction, because the protein data is non-linear high dimensional sequence data and also have missing features value. As we know that the SVM can be used for non-liner high dimensional data, the C5.0 can be used for all possible data including missing value and the Bayesian is suitable for sequence data. As a result, the above discussed classification technique is more suited in case of protein data classification and prediction. But our experimental analysis highlights that the C5.0 gives highest value for all class in all models for every measures. Therefore, based on the our experimental analysis we can conclude that the C5.0 is more suited in protein classification.  The above discussion has not found a way to know the exact classification technique for particular class. To do this, we have made a comparative analysis of performance of all classification technique for all classes with respect to accuracy, precision, recall, f-measures, MCC, sensitivity and specificity. The comparative result shown in figure \ref{fig:a} to figure \ref{fig:g}. 

\begin{figure}[!hbt]
	\vspace{-0.5cm}
	\centering
		\includegraphics[width=1\linewidth,height=2.5in]{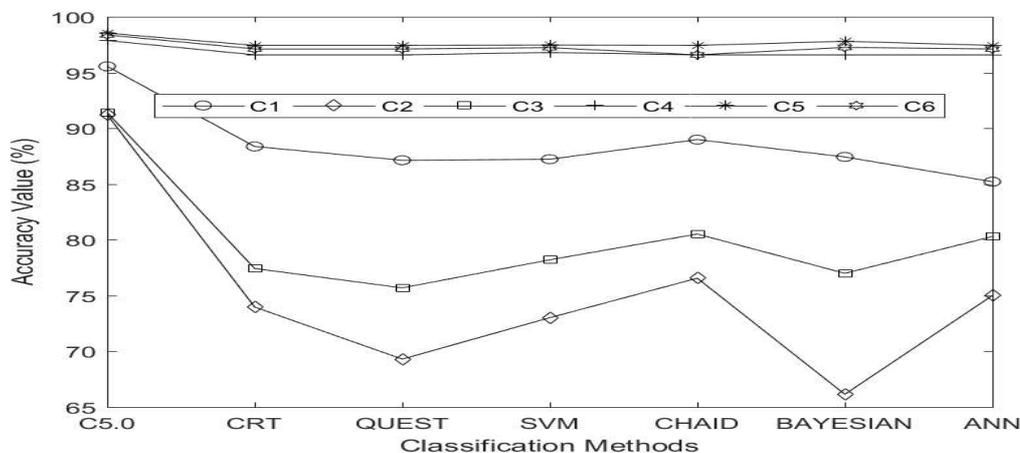}
		\caption{Comparative analysis of accuracy for all models and classes}
		\label{fig:a}
		\vspace{-0.25cm}
\end{figure}
Here it can be clearly seen that the accuracy for class C3, C4, C5 and C6 are very skew and have high accuracy for all model rather than the class C1, C2 and C3. The Bayesian based model give less accuracy than other models for all classes. The accuracy of C5.0 model is high for all classes. 

\begin{figure}[!hbt]
	\centering
	\includegraphics[width=1\linewidth,height=3in]{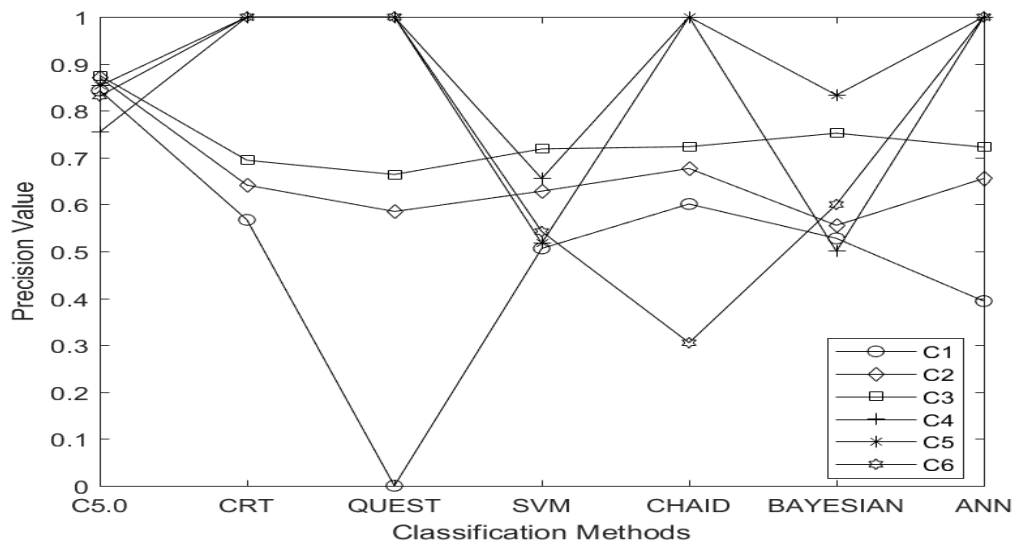}
	\caption{Comparative analysis of Precision for all models and classes}
	\label{fig:d}
\end{figure}
In figure \ref{fig:d}, the value at point 1 indicates the "\#DIV/0!". The "\#DIV/0!" indicates that the models are under-fitted in those class.  Here it can be seen that the C5.0 models gives high precision value for all classes and the QUEST gives very low precision value for class C1. The high precision value indicates that the C5.0 gives a balanced result in protein classification. 

\begin{figure}[!hbt]
	\centering
	\includegraphics[width=1\linewidth,height=3in]{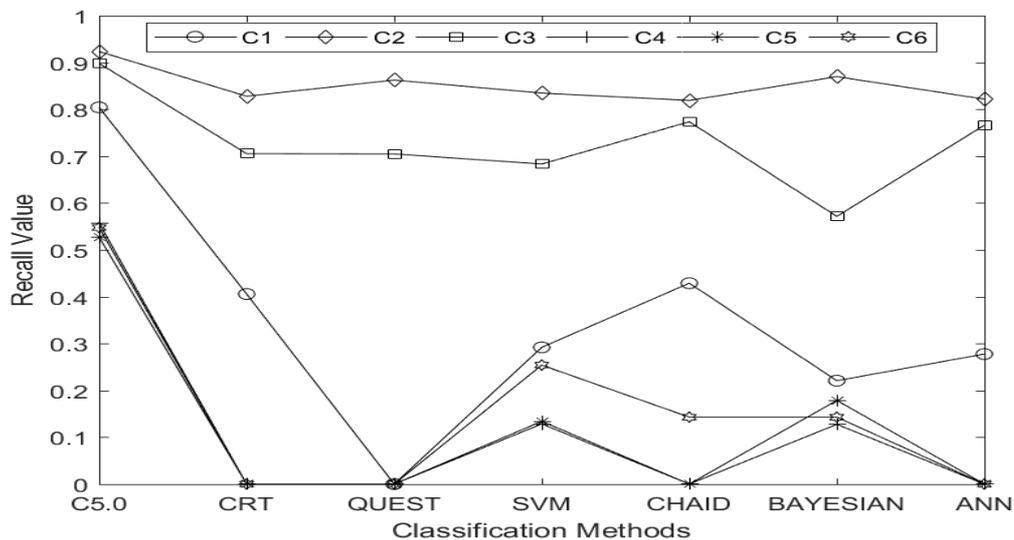}
	\caption{Comparative analysis of Recall for all models and classes}
	\label{fig:e}
\end{figure}

Form figure \ref{fig:e}, it is clearly seen that the recall value is high for class C2 and C3 with respect to other class and the C5.0 model gives high recall value for every class of data. On the other hand the CRT gives lower value for class C4, C5 and C6, Quest gives lower value for C1, C4, C5 and C6 class of data, CHAID gives lower value for C4 and C5 class of data and ANN gives lower recall value for C4, C5 and C6 class of data.

\begin{figure}[!hbt]
		\centering
		\includegraphics[width=1\linewidth,height=3in]{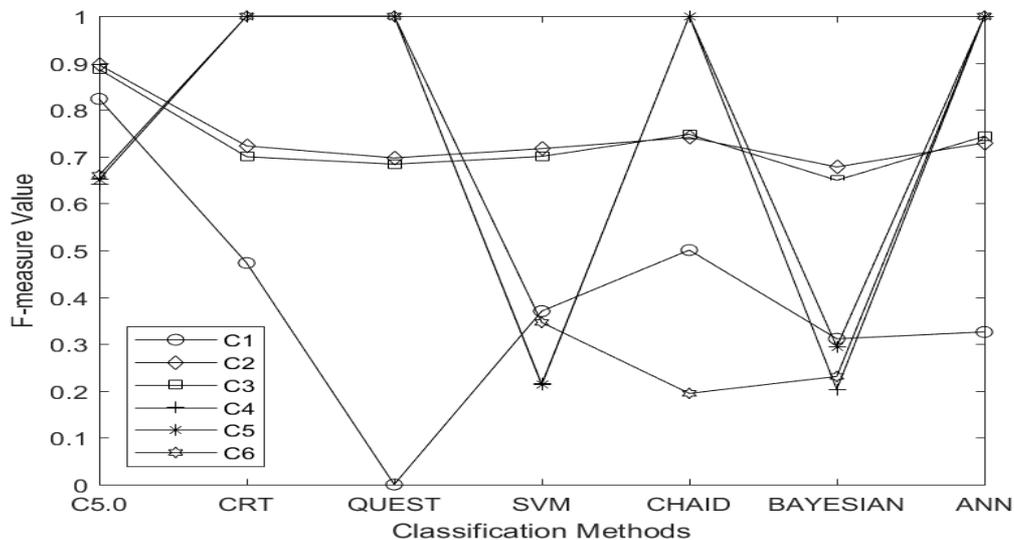}
		\caption{Comparative analysis of F-measure for all models and classes}
		\label{fig:b}
\end{figure}
In figure\ref{fig:b}, the value at point 1 for class indicate the "\#DIV/0!". Here it can be clearly seen that the f-measure value for all class is high in C5.0 and have very less margin for class C2 and C3 in all model. The quest produce very less f-measure value for class C1, that is 0. 

\begin{figure}[!hbt]
		\centering
		\includegraphics[width=1\linewidth,height=3in]{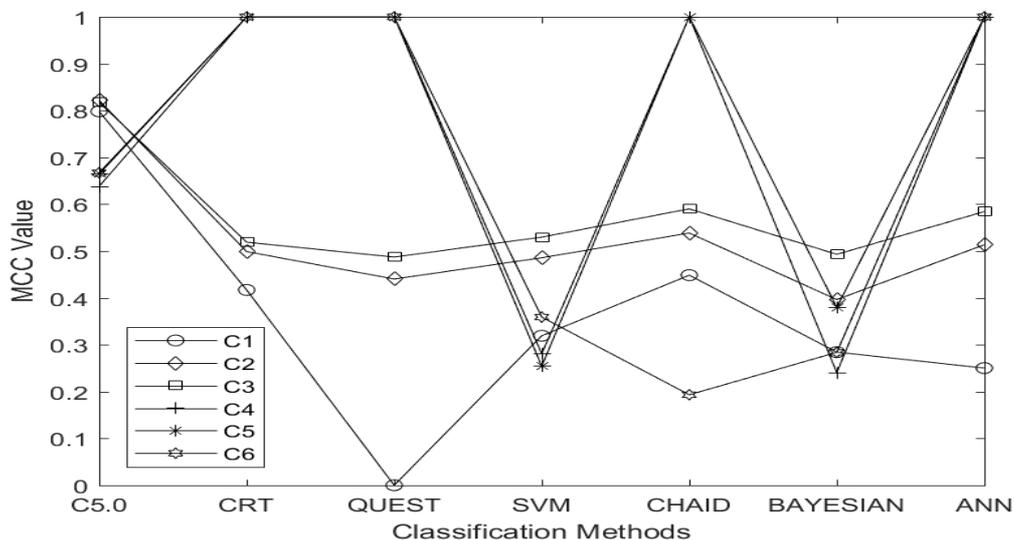}
		\caption{Comparative analysis of MCC for all models and classes}
		\label{fig:c}
\end{figure}
The value at point 1 indicates that the "\#DIV/0!" in figure \ref{fig:c}. The C5.0 based model gives highest MCC value for all class and the QUEST based Model gives lowest value for class C1 i e. 0. Here it can be clearly seen that the MCC value for class C2 and C3 in all models have very less variations. Finally, we can conclude that, if we have consider MCC as a evaluation metrics then C5.0 is more suited in classification rather than others.

\begin{figure}[!hbt]
	\centering
	\includegraphics[width=1\linewidth,height=3in]{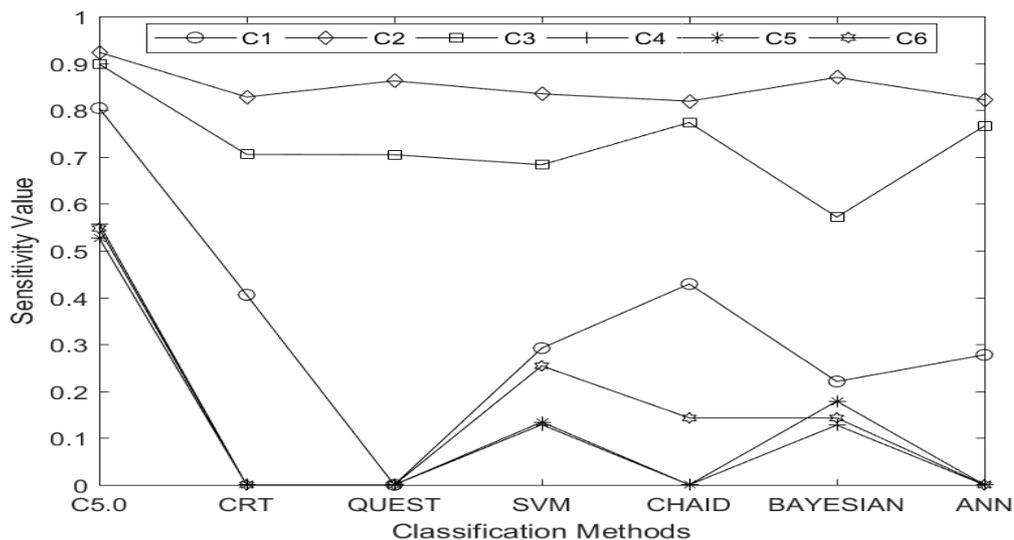}
	\caption{Comparative analysis of Sensitivity for all models and classes}
	\label{fig:f}
\end{figure}
The result of the sensitivity is almost similar to the recall value. Lower the sensitivity higher the precision value. The lower sensitivity value may give a highest selection value. As similar to the other measures the C5.0 model gives high sensitivity value than others.  
\begin{figure}[!hbt]
	
	\centering
	\includegraphics[width=1\linewidth,height=3in]{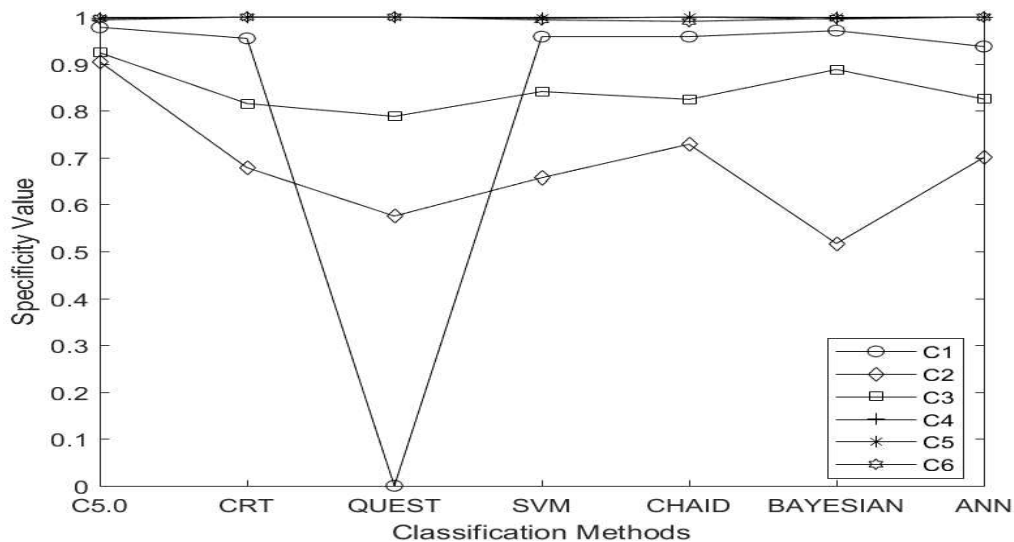}
	\caption{Comparative analysis of Specificity for all models and classes}
	\label{fig:g}
\end{figure}

In figure \ref{fig:g}, it can be clearly seen that the half of the class gives "\#DIV/0!" value for all models. All these class data are under-fitted in this model. In this measure the C5.0 gives high specificity value than others and the QUEST give lower specificity value for class C1.

Finally, we can conclude that the Model C5.0 gives better results than the other measures. The highest performance value for all classes and the models are shown in Table \ref{tab:Highest performance of all model for all class}.\\\\

\begin{center}
	\begin{table}[h!]
		\renewcommand{\arraystretch}{1.2}
		\centering
		\caption{Highest performance of all model for all class}
		\label{tab:Highest performance of all model for all class}
		
		\begin{tabular}{llllllll}		\hline
			\textbf{}	&	\textbf{C5.0}	&	\textbf{CRT}&\textbf{QUEST} &\textbf{SVM} & \textbf{CHAID} &\textbf{BAYESIAN}&\textbf{ANN}	\\\hline
			Accuracy	&	All Class	&	 -	&	- 	&	 -	&	- 	&	- 	&	- 	\\
			Sensitivity	&	All Class	&	 -	&	- 	&	 -	&	- 	&	- 	&	- 	\\
			Specificity	&	C1, C2, C3	&	C4, C5	&	C4, C5	&	 -	&	C4	&	- 	&	C4, C5	\\
			Precision	&	All Class	&	 -	&	- 	&	 -	&	- 	&	- 	&	- 	\\
			Recall	&	All Class	&	 -	&	- 	&	 -	&	- 	&	- 	&	- 	\\
			F-measure	&	All Class	&	 -	&	- 	&	 -	&	- 	&	- 	&	- 	\\
			MCC	&	All Class	&	 -	&	- 	&	 -	&	- 	&	- 	&	- 	\\

			\hline
		\end{tabular}
	\end{table}
\end{center}
This table indicates that the C5.0 classification technique gives better result than others classification technique on my data set. Here it can be clearly seen that the all performance measures except Specificity gives better result, when we use C5.0 classification technique. The specificity of class C1, C2 and C3 are best in C5.0 and rest of the class specificity gives best result in CRT, QUEST, CHAID and Neural network-based model. The specificity of class C3, C4 and C5 is low in C5.0 classification model because the data of class C3, C4 and C5 are more imbalanced than C1, C2 and C3. Finally, we can say that C5.0 classification technique performs better in our dataset. To prove this, we have performed another experimental analysis that cover all data as a single data set (not in class) using SPSS tool. The result of the analysis is shown in Table \ref{tab: Overall performance table of all models}
\begin{center}
	\begin{table}[h!]
		\renewcommand{\arraystretch}{1.2}
		\centering
		\caption{Overall performance table of all models}
		\label{tab: Overall performance table of all models}
		
		\begin{tabular}{lllllll}		\hline
			\textbf{S.No.}	&	\textbf{Model}	&  \multicolumn{2}{c}{	\textbf{Data}}&  \multicolumn{2}{c}{\textbf{Accuracy}} &\textbf{Elapsed time} \\\hline
			&		&	Correct	&	Wrong	&	Correct	&	Wrong	&		\\
		1	&	CRT 	&	2,925	&	1,443	&	66.96\%	&	33.04\%	&	 0 hours, 0 mins, 40 secs	\\
		2	&	QUEST	&	2,692	&	1,676	&	61.63\%	&	38.37\%	&	0 hours, 0 mins, 1 secs	\\
		3	&	CHAID	&	2,985	&	1,383	&	68.34\%	&	31.66\%	&	0 hours, 0 mins, 5 secs	\\
		4	&	C5.0	&	3,778	&	590	&	86.49\%	&	13.51\%	&	0 hours, 0 mins, 2 secs	\\
		5	&	ANN	&	2,900	&	1,468	&	66.39\%	&	33.61\%	&	0 hours, 0 mins, 2 secs	\\
		6	&	SVM	&	2,837	&	1,531	&	64.95\%	&	35.05\%	&	0 hours, 0 mins, 39 secs	\\
		7	&	BAYSEAN	&	2,669	&	1,699	&	61.10\%	&	38.90\%	&	0 hours, 0 mins, 3 secs	\\

			\hline
		\end{tabular}
	\end{table}
\end{center}
This table gives a brief summary of different classification technique including the prediction of total correct and wrong data, accuracy of correct and wrong data and total elapsed time that has been taken by machine to predict and produce the result.  From Table \ref{tab: Overall performance table of all models}, one can clearly seen that the C5.0 classification technique classifies total 3778 out of 4368 sample as a correct and very less total 590 samples as wrong. The overall accuracy of C5.0 classification technique on this data set is 86.49 and also takes very less amount of computation time i.e. 2 sec. From the above discussion, it can be concluded that the C5.0 classification technique performs better than the other classification technique. So, finally, we can say that C5.0 classification technique can be use as a classification technique to classify the protein data and these 37 features (shown in Table \ref{tab:List of features selected by different models with their importance factor}) that are classified by C5.0 can be used for function prediction. 

\section{Conclusions}
Many previous researcheshave been conducted in the field of computational biology and determine the meaningful and accurate features for protein function prediction. Here we discuss seven different type of classification techniques such as CRT, QUEST, CHAID, C5.0, ANN, SVM and BAYSEAN for protein feature classification and predictions. For classification and predictions, we have conducted an experimental analysis on 4368 numbers of sample data of proteins of human category. The experimental results highlight the properties of different classification techniques and found that class C4, C5 and C6 data are more imbalanced than the others class of data. The imbalanced data affect the performance of classification techniques, as a result the precision and recall value of these classes data are relatively very low than others classes.The experimental result shown here suggest that the C5.0 classification technique can be used for classification and prediction of protein based on features, and the features that are classified can be used in function prediction.

\bibliographystyle{apalike}      
\bibliography{reportnew}

\begin{thebibliography}{}

\bibitem[Amidi et~al., 2016]{21}
Amidi, A., Amidi, S., Vlachakis, D., Paragios, N., and Zacharaki, E.~I. (2016).
\newblock A machine learning methodology for enzyme functional classification
  combining structural and protein sequence descriptors.
\newblock In {\em International Conference on Bioinformatics and Biomedical
  Engineering}, pages 728--738. Springer.

\bibitem[Baldi et~al., 2000]{38}
Baldi, P., Brunak, S., Chauvin, Y., Andersen, C.~A., and Nielsen, H. (2000).
\newblock Assessing the accuracy of prediction algorithms for classification:
  an overview.
\newblock {\em Bioinformatics}, 16(5):412--424.

\bibitem[Borro et~al., 2006]{17}
Borro, L.~C., Oliveira, S.~R., Yamagishi, M.~E., Mancini, A.~L., Jardine,
  J.~G., Mazoni, I., Santos, E.~D., Higa, R.~H., Kuser, P.~R., and Neshich, G.
  (2006).
\newblock Predicting enzyme class from protein structure using bayesian
  classification.
\newblock {\em Genet. Mol. Res}, 5(1):193--202.

\bibitem[Caruana and Sa, 2003]{24}
Caruana, R. and Sa, V. R.~d. (2003).
\newblock Benefitting from the variables that variable selection discards.
\newblock {\em Journal of machine learning research}, 3(Mar):1245--1264.

\bibitem[Cheeseman et~al., 1988]{36}
Cheeseman, P., Self, M., Kelly, J., Taylor, W., Freeman, D., and Stutz, J.
  (1988).
\newblock Bayesian classification$\backslash$in aaai 88.
\newblock In {\em The 7th National Conference on Artificial Intelligence},
  volume 607, page 611.

\bibitem[Das et~al., 2015]{2}
Das, S., Sillitoe, I., Lee, D., Lees, J.~G., Dawson, N.~L., Ward, J., and
  Orengo, C.~A. (2015).
\newblock Cath funfhmmer web server: protein functional annotations using
  functional family assignments.
\newblock {\em Nucleic acids research}, 43(W1):W148--W153.

\bibitem[Dobson and Doig, 2005]{16}
Dobson, P.~D. and Doig, A.~J. (2005).
\newblock Predicting enzyme class from protein structure without alignments.
\newblock {\em Journal of molecular biology}, 345(1):187--199.

\bibitem[Garg and Raghava, 2008]{13}
Garg, A. and Raghava, G.~P. (2008).
\newblock A machine learning based method for the prediction of secretory
  proteins using amino acid composition, their order and similarity-search.
\newblock {\em In silico biology}, 8(2):129--140.

\bibitem[Haijun et~al., 2007]{33}
Haijun, X., Fang, P., Ling, W., and Hongwei, L. (2007).
\newblock Ad hoc-based feature selection and support vector machine classifier
  for intrusion detection.
\newblock In {\em Grey Systems and Intelligent Services, 2007. GSIS 2007. IEEE
  International Conference on}, pages 1117--1121. IEEE.

\bibitem[Hall and Holmes, 2003]{25}
Hall, M.~A. and Holmes, G. (2003).
\newblock Benchmarking attribute selection techniques for discrete class data
  mining.
\newblock {\em IEEE Transactions on Knowledge and Data engineering},
  15(6):1437--1447.

\bibitem[Han et~al., 2011]{37}
Han, J., Pei, J., and Kamber, M. (2011).
\newblock {\em Data mining: concepts and techniques}.
\newblock Elsevier.

\bibitem[Jackson and Bartek, 2009]{3}
Jackson, S.~P. and Bartek, J. (2009).
\newblock The dna-damage response in human biology and disease.
\newblock {\em Nature}, 461(7267):1071.

\bibitem[Jensen et~al., 2002]{15}
Jensen, L.~J., Skovgaard, M., and Brunak, S. (2002).
\newblock Prediction of novel archaeal enzymes from sequence-derived features.
\newblock {\em Protein Science}, 11(12):2894--2898.

\bibitem[Karp, 1998]{23}
Karp, P.~D. (1998).
\newblock What we do not know about sequence analysis and sequence databases.
\newblock {\em Bioinformatics (Oxford, England)}, 14(9):753--754.

\bibitem[Karunapala, 2015]{1}
Karunapala, E. (2015).
\newblock {\em Protein Function Prediction Using Machine Learning}.
\newblock PhD thesis.

\bibitem[Kotlyar et~al., 2014]{8}
Kotlyar, M., Pastrello, C., Pivetta, F., Sardo, A.~L., Cumbaa, C., Li, H.,
  Naranian, T., Niu, Y., Ding, Z., Vafaee, F., et~al. (2014).
\newblock In silico prediction of physical protein interactions and
  characterization of interactome orphans.
\newblock {\em Nature methods}, 12(1):79.

\bibitem[Lee et~al., 2008]{11}
Lee, B.~J., Lee, H.~G., and Ryu, K.~H. (2008).
\newblock Design of a novel protein feature and enzyme function classification.
\newblock In {\em Computer and Information Technology Workshops, 2008. CIT
  Workshops 2008. IEEE 8th International Conference on}, pages 450--455. IEEE.

\bibitem[Li et~al., 2016]{9}
Li, Y.~H., Xu, J.~Y., Tao, L., Li, X.~F., Li, S., Zeng, X., Chen, S.~Y., Zhang,
  P., Qin, C., Zhang, C., et~al. (2016).
\newblock Svm-prot 2016: A web-server for machine learning prediction of
  protein functional families from sequence irrespective of similarity.
\newblock {\em PloS one}, 11(8):e0155290.

\bibitem[Li et~al., 2017]{5a}
Li, Y.~H., Yu, C.~Y., Li, X.~X., Zhang, P., Tang, J., Yang, Q., Fu, T., Zhang,
  X., Cui, X., Tu, G., et~al. (2017).
\newblock Therapeutic target database update 2018: enriched resource for
  facilitating bench-to-clinic research of targeted therapeutics.
\newblock {\em Nucleic acids research}, 46(D1):D1121--D1127.

\bibitem[Liang and Bose, 1996]{30}
Liang, P. and Bose, N. (1996).
\newblock Neural network fundamentals with graphs, algorithms, and
  applications.
\newblock {\em McGraw-Hiil, New York}.

\bibitem[Lin et~al., 2011]{19}
Lin, W.-Z., Fang, J.-A., Xiao, X., and Chou, K.-C. (2011).
\newblock idna-prot: identification of dna binding proteins using random forest
  with grey model.
\newblock {\em PloS one}, 6(9):e24756.

\bibitem[Matthews, 1975]{39}
Matthews, B.~W. (1975).
\newblock Comparison of the predicted and observed secondary structure of t4
  phage lysozyme.
\newblock {\em Biochimica et Biophysica Acta (BBA)-Protein Structure},
  405(2):442--451.

\bibitem[Mer and Andrade-Navarro, 2013]{14}
Mer, A.~S. and Andrade-Navarro, M.~A. (2013).
\newblock A novel approach for protein subcellular location prediction using
  amino acid exposure.
\newblock {\em BMC bioinformatics}, 14(1):342.

\bibitem[Milanovi{\'c} and Stamenkovi{\'c}, 2016]{28}
Milanovi{\'c}, M. and Stamenkovi{\'c}, M. (2016).
\newblock Chaid decision tree: Methodological frame and application.
\newblock {\em Economic Themes}, 54(4):563--586.

\bibitem[Nizar et~al., 2008]{32}
Nizar, A., Dong, Z., and Wang, Y. (2008).
\newblock Power utility nontechnical loss analysis with extreme learning
  machine method.
\newblock {\em IEEE Transactions on Power Systems}, 23(3):946--955.

\bibitem[Pang and Gong, 2009]{29}
Pang, S.-l. and Gong, J.-z. (2009).
\newblock C5. 0 classification algorithm and application on individual credit
  evaluation of banks.
\newblock {\em Systems Engineering-Theory \& Practice}, 29(12):94--104.

\bibitem[Piovesan et~al., 2015]{6}
Piovesan, D., Giollo, M., Leonardi, E., Ferrari, C., and Tosatto, S.~C. (2015).
\newblock Inga: protein function prediction combining interaction networks,
  domain assignments and sequence similarity.
\newblock {\em Nucleic acids research}, 43(W1):W134--W140.

\bibitem[Poux et~al., 2017]{22}
Poux, S., Arighi, C.~N., Magrane, M., Bateman, A., Wei, C.-H., Lu, Z., Boutet,
  E., Bye-A-Jee, H., Famiglietti, M.~L., Roechert, B., et~al. (2017).
\newblock On expert curation and scalability: Uniprotkb/swiss-prot as a case
  study.
\newblock {\em Bioinformatics}, 33(21):3454--3460.

\bibitem[Rentzsch and Orengo, 2013]{7}
Rentzsch, R. and Orengo, C.~A. (2013).
\newblock Protein function prediction using domain families.
\newblock In {\em BMC bioinformatics}, volume~14, page~S5. BioMed Central.

\bibitem[Samb et~al., 2012]{34}
Samb, M.~L., Camara, F., Ndiaye, S., Slimani, Y., and Esseghir, M.~A. (2012).
\newblock A novel rfe-svm-based feature selection approach for classification.
\newblock {\em International Journal of Advanced Science and Technology},
  43(1):27--36.

\bibitem[Schwartz et~al., 2011]{26}
Schwartz, C.~E., Sprangers, M.~A., Oort, F.~J., Ahmed, S., Bode, R., Li, Y.,
  and Vollmer, T. (2011).
\newblock Response shift in patients with multiple sclerosis: an application of
  three statistical techniques.
\newblock {\em Quality of Life Research}, 20(10):1561--1572.

\bibitem[Singh and Tripathi, 2016]{10}
Singh, U. and Tripathi, S. (2016).
\newblock Protein classification using hybrid feature selection technique.
\newblock In {\em International Conference on Smart Trends for Information
  Technology and Computer Communications}, pages 813--821. Springer.

\bibitem[Tiwari and Srivastava, 2014]{35}
Tiwari, A.~K. and Srivastava, R. (2014).
\newblock A survey of computational intelligence techniques in protein function
  prediction.
\newblock {\em International journal of proteomics}, 2014.

\bibitem[Vapnik, 2013]{31}
Vapnik, V. (2013).
\newblock {\em The nature of statistical learning theory}.
\newblock Springer science \& business media.

\bibitem[Weinberg and Chandel, 2015]{4}
Weinberg, S.~E. and Chandel, N.~S. (2015).
\newblock Targeting mitochondria metabolism for cancer therapy.
\newblock {\em Nature chemical biology}, 11(1):9.

\bibitem[Wu et~al., 2008]{20}
Wu, J., Liu, H., Duan, X., Ding, Y., Wu, H., Bai, Y., and Sun, X. (2008).
\newblock Prediction of dna-binding residues in proteins from amino acid
  sequences using a random forest model with a hybrid feature.
\newblock {\em Bioinformatics}, 25(1):30--35.

\bibitem[Yadav and Jayaraman, 2012]{12}
Yadav, A. and Jayaraman, V.~K. (2012).
\newblock Structure based function prediction of proteins using fragment
  library frequency vectors.
\newblock {\em Bioinformation}, 8(19):953.

\bibitem[Yadav et~al., 2015]{18}
Yadav, S.~K., Bhola, A., and Tiwari, A.~K. (2015).
\newblock Classification of enzyme functional classes and subclasses using
  support vector machine.
\newblock In {\em Futuristic Trends on Computational Analysis and Knowledge
  Management (ABLAZE), 2015 International Conference on}, pages 411--417. IEEE.

\bibitem[Yang et~al., 2015]{5}
Yang, H., Qin, C., Li, Y.~H., Tao, L., Zhou, J., Yu, C.~Y., Xu, F., Chen, Z.,
  Zhu, F., and Chen, Y.~Z. (2015).
\newblock Therapeutic target database update 2016: enriched resource for bench
  to clinical drug target and targeted pathway information.
\newblock {\em Nucleic acids research}, 44(D1):D1069--D1074.

\bibitem[Zwartjes et~al., 2016]{27}
Zwartjes, A., Havinga, P.~J., Smit, G.~J., and Hurink, J.~L. (2016).
\newblock Quest: Eliminating online supervised learning for efficient
  classification algorithms.
\newblock {\em Sensors}, 16(10):1629.

\end{thebibliography}

\end{document}